%% file: main.tex
\documentclass{article}

\usepackage[preprint]{neurips_2025}

\usepackage[T1]{fontenc}
\usepackage{amsmath,amssymb,amsfonts}
\usepackage{booktabs}
\usepackage{graphicx}
\usepackage{xspace}
\usepackage{xcolor}
\usepackage[colorlinks=true,linkcolor=black,citecolor=black,urlcolor=black]{hyperref}
\usepackage{microtype}
\usepackage{placeins}
\usepackage{subcaption}
\usepackage{multirow}
\usepackage{tikz}
\usetikzlibrary{positioning, arrows.meta, calc, decorations.pathreplacing}
\usepackage{pgfplots}
\pgfplotsset{compat=1.18}
\usepgfplotslibrary{fillbetween}
\usetikzlibrary{patterns}
\usepackage[ruled,vlined,linesnumbered]{algorithm2e}

\pgfplotsset{
  paperplot/.style={
    axis x line*=bottom,
    axis y line*=left,
    axis line style={black!60, line width=0.45pt},
    every axis/.append style={line width=0.35pt},
    tick style={line width=0.3pt, color=black!45},
    tick align=outside,
    every tick label/.append style={font=\footnotesize},
    label style={font=\small},
    grid=none,
    clip=false,
  },
}

\definecolor{s0color}{HTML}{009988}   
\definecolor{s0lora}{HTML}{EE7733}    
\definecolor{s0offset}{HTML}{CC3311}  

\input{defs}

\title{%
  \methodshort Tuning: Zero-Overhead Adaptation of Hybrid Recurrent-Attention Models%
}

\author{%
  Jack Young \\
  Indiana University \\
  \texttt{youngjh@iu.edu} \\
}

\begin{document}

\maketitle

\begin{abstract}
Using roughly 48 execution-verified HumanEval training solutions, tuning a
single initial state matrix per recurrent layer, with zero inference
overhead, outperforms LoRA by $\loragap$~\pp{} ($p < 0.001$) on HumanEval.
The method, which we call \method, optimizes one state matrix per recurrent
layer while freezing all model weights.
On Qwen3.5-4B (\gdn hybrid), \method improves greedy pass@1 by
$\mainresult \pm 1.7$~\pp{} (10 seeds).
On \falcon-7B (Mamba-2 hybrid), \methodshort reaches $71.8\% \pm 1.3$ and
LoRA reaches $71.4\% \pm 2.4$ (3~seeds), statistically indistinguishable at this sample size while requiring no weight merging.
Cross-domain transfer is significant on MATH-500 ($+4.8$~\pp, $p = 0.00002$, 8 seeds) and GSM8K ($+2.8$~\pp, $p = 0.0003$, 10 seeds); a text-to-SQL benchmark (Spider) shows no transfer, consistent with the trajectory-steering mechanism.
A prefix-tuning control on a pure Transformer (Qwen2.5-3B) \emph{degrades}
performance by $-13.9$~\pp{} under all nine configurations tested.
On Qwen3.5, a per-step state-offset variant reaches $+27.1$~\pp{}, above both
\methodshort and LoRA but with per-step inference cost. Taken together, the
results show that recurrent state initialization is a strong
zero-inference-overhead PEFT surface for hybrid language models when verified
supervision is scarce.
The tuned state is a ${\sim}$48\,MB file; task switching requires no weight merging or model reload.
Code and library: \url{https://github.com/jackyoung27/s0-tuning}.
\end{abstract}

\section{Introduction}
\label{sec:intro}

Production language models increasingly combine recurrent layers with standard
attention in a single backbone~\cite{gated_deltanet,mamba2,falconh1}.
The recurrent component varies.
Qwen3.5 interleaves three \gdn layers per attention layer; \falcon places
Mamba-2 and attention heads in parallel within each layer.
These hybrid architectures achieve subquadratic sequence cost while retaining
the in-context learning of full attention. They also expose a new adaptation
surface that pure Transformers lack: the recurrent hidden state $\state$, a
matrix updated at every token.

LoRA~\cite{lora} and its descendants adapt weight matrices, a strategy introduced
for and validated on Transformers.
Hybrid models, however, carry a per-layer state matrix that accumulates distributional
information across the entire context window and is set to zero by default.
Replacing that zero with a learned value steers the model toward a target task.

Hybrid recurrent-attention backbones now ship in major open-weight releases~\cite{gated_deltanet,falconh1,mamba3}, yet PEFT has not kept pace: LoRA and prefix tuning target weight matrices and leave the recurrent state at its default zero.
\method optimizes a single state matrix $\svec$ per recurrent layer (12.6M
parameters, 0.3\% of Qwen3.5-4B) with all weights frozen.
On HumanEval, this improves greedy pass@1 by $\mainresult \pm 1.7$~\pp,
outperforming LoRA by $\loragap$~\pp ($p < 0.001$, 10 seeds each) with zero
inference overhead.
On \falcon-7B, a Mamba-2 hybrid with a fundamentally different recurrence
family, \methodshort reaches $71.8\% \pm 1.3$ and LoRA reaches $71.4\% \pm 2.4$ (3 seeds each), statistically indistinguishable at this sample size.
Prefix tuning applied to a pure Transformer (Qwen2.5-3B) degrades performance
by $-13.9$~\pp{} under all nine configurations tested.

This is a small-data paper. Roughly 48 verified HumanEval solutions move
Qwen3.5-4B by $\mainresult$~\pp{} and beat LoRA by $\loragap$~\pp{} over
10 seeds; Falcon supplies a 3-seed second architecture at $71.8\%$ versus
$71.4\%$ for LoRA, and the math benchmarks add smaller but significant gains.
That is enough to establish the main point: recurrent state is a high-leverage
adaptation surface for hybrid models, while the remaining experiments map
transfer, scale, and failure modes.

How does a perturbation to a single initial matrix propagate to such large gains?
$\svec$'s direct influence on output logits decays to 0.03\% KL ratio by the
end of the prompt, yet 23 of 27 FAIL-to-PASS flips diverge from the baseline
at the very first generated character (85\%).
The initial perturbation is amplified by the recurrence into a qualitatively
different generation trajectory. This is trajectory-steering, distinct from the uniform
weight modification that LoRA applies.

\paragraph{Contributions.}
\begin{itemize}
  \item \method: a zero-inference-overhead PEFT method that outperforms LoRA by $\loragap$~\pp{} on \gdn{} (Qwen3.5-4B, $p < 0.001$) and is statistically indistinguishable from LoRA on Mamba-2 in a 3-seed FalconH1 comparison ($71.8\%$ vs.\ $71.4\%$).
  \item Parameter-matched control: LoRA at rank~64 (12.6M, equal to \methodshort) degrades by $-15.5$~\pp{} in this small-data regime, arguing against parameter count alone as the explanation.
  \item Transformer negative control: prefix tuning on a pure Transformer (Qwen2.5-3B) degrades performance by $-13.9$~\pp{} under all nine configurations.
  \item Mechanistic analysis: $\svec$'s direct influence decays to 0.03\% KL by the end of the prompt, yet 85\% of corrected solutions diverge at the first generated character.
  \item Scaling and pass@$k$: a state-offset variant achieves $+27.1$~\pp; pass@10 reaches 88.5\% versus 66.7\% for LoRA in a separate 3-seed sampled evaluation.
\end{itemize}

\begin{figure}[t]
  \centering
  \includegraphics[width=\textwidth]{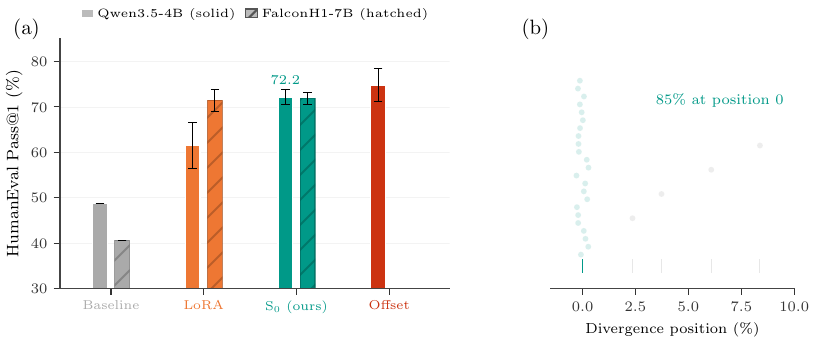}
  \caption{%
    \textbf{Overview of \method.}
    (a)~Cross-architecture comparison on HumanEval: \methodshort (teal) outperforms
    LoRA on Qwen3.5-4B and is tied with it on \falcon-7B (hatched bars). The
    Qwen comparison is significant ($p < 0.001$; ***); the Falcon 3-seed
    comparison is statistically indistinguishable from LoRA.
    Offset is shown for Qwen only (not applicable to \falcon's recurrence).
    Error bars show $\pm$1 std across seeds.
    (b)~First-character divergence: of 27 FAIL-to-PASS flips, 23 (85\%) diverge from
    baseline at the very first generated character (teal dots at position 0).
  }
  \label{fig:hero}
\end{figure}

\section{Background}
\label{sec:background}

\subsection{Recurrent State Mechanics}
\label{sec:bg_state}

In \gdn~\cite{gated_deltanet}, each recurrent layer maintains a state matrix
$\state \in \stateshape$ updated via the gated delta rule:
\begin{equation}
  \state = \alpha_t\, S_{t-1}\!\left(I - \beta_t\, k_t k_t^\top\right) + \beta_t\, v_t k_t^\top,
  \label{eq:gdn_update}
\end{equation}
where $\alpha_t \in (0,1)$ is a decay gate, $\beta_t \in (0,1)$ controls write
strength, and $k_t$, $v_t$ are key and value vectors.
The term $(I - \beta_t k_t k_t^\top)$ erases the old association for key~$k_t$
before writing the new one.
Mamba-2~\cite{mamba2} uses a structurally similar update through its structured
state-space duality (SSD):
\begin{equation}
  \state = \bar{A}_t\, S_{t-1} + \bar{B}_t\, x_t,
  \label{eq:ssd_update}
\end{equation}
where $\bar{A}_t$ is a scalar gate and $\bar{B}_t$ an input projection.
In both architectures, the state is a full matrix: a single \gdn layer in
Qwen3.5-4B carries ${\sim}$524K state entries ($H{=}32, K{=}V{=}128$).
Mamba-1~\cite{mamba}, by contrast, uses a diagonal state where each of $N{=}16$
dimensions evolves independently, with no cross-feature interactions.
This structural gap matters.
A full matrix state has the capacity to encode cross-feature correlations that a diagonal state lacks, consistent with a state-expressiveness threshold below which initial-state tuning becomes ineffective.
By default, both architectures initialize $S_0 = 0$; our method replaces this
zero initialization with a learned value.

\section{Method}
\label{sec:method}

\subsection{\methodshort Tuning}
\label{sec:method_s0}

Hybrid recurrent-attention models expose a recurrent state at every recurrent
layer; we treat the initial value of that state as the adaptation surface. For
each recurrent layer $\ell$, we introduce a learnable tensor
$\svec^{(\ell)}$ with the same shape as the layer's native recurrent state and
freeze all backbone parameters $\theta$. Given a prompt-completion pair
$(x, y)$, the recurrent layer now starts from $\alpha \svec^{(\ell)}$ rather
than the default zero state, where $\alpha$ is a scalar state-scaling
hyperparameter chosen per architecture. We optimize only the collection
$\{\svec^{(\ell)}\}_{\ell=1}^L$ using a completion-only objective,
\[
\mathcal{L}(\svec) =
\frac{1}{N}\sum_{i=1}^N
\mathrm{CE}\!\left(y_i^{\mathrm{comp}},
f_\theta(x_i; \alpha \svec)\right) +
\lambda \sum_{\ell=1}^L \|\svec^{(\ell)}\|_2^2,
\]
where prompt tokens are masked so gradients act only on the target completion.
\begin{enumerate}
  \item Initialize $\svec^{(\ell)} = 0$ for each recurrent layer $\ell$.
  \item Inject $\svec^{(\ell)}$ as the initial hidden state before the first token.
  \item Optimize $\svec^{(\ell)}$ via gradient descent on a completion-only loss over correct solutions, with all model weights frozen.
  \item At inference, $\svec$ is absorbed into the recurrent state at $t=1$ and adds zero cost at any subsequent timestep.
\end{enumerate}

\begin{algorithm}[t]
\caption{\method}
\label{alg:s0_tuning}
\KwIn{Hybrid model $f_\theta$, correct solutions $\{(x_i, y_i)\}_{i=1}^N$, scaling $\alpha$}
\KwOut{Learned initial states $\{\svec^{(\ell)}\}_{\ell=1}^L$}
Initialize $\svec^{(\ell)} \gets 0$ for each recurrent layer $\ell$\;
\For{step $= 1$ \KwTo $T$}{
  Set initial state $S_0^{(\ell)} \gets \alpha \cdot \svec^{(\ell)}$ for each recurrent layer\;
  Compute loss $\mathcal{L} = \frac{1}{N}\sum_i \mathrm{CE}(y_i, f_\theta(x_i;\, \alpha\svec)) + \lambda\sum_\ell\|\svec^{(\ell)}\|_2^2$\;
  Update $\svec^{(\ell)} \gets \svec^{(\ell)} - \eta \nabla_{\svec^{(\ell)}} \mathcal{L}$\;
}
\Return $\{\svec^{(\ell)}\}_{\ell=1}^L$\;
\end{algorithm}

Qwen3.5 uses $\alpha = 0.07$; \falcon uses $\alpha = 0.65$ (Section~\ref{sec:results_alpha}).
All other hyperparameters are given in Section~\ref{sec:setup_stats}.

We also evaluate a state-offset variant~\cite{state_offset} that adds a learned tensor $\Delta S^{(\ell)}$ at every timestep (non-zero overhead); see Section~\ref{sec:results_offset}.

\paragraph{Zero-overhead property.}
Because $\svec$ is injected only at $t{=}0$, it is absorbed into the running state at $t{=}1$; every subsequent step executes the unmodified recurrence with no adapter branch and no weight merging.
\methodshort is therefore zero-inference-overhead as a structural property of the computation graph, not as an empirical approximation.

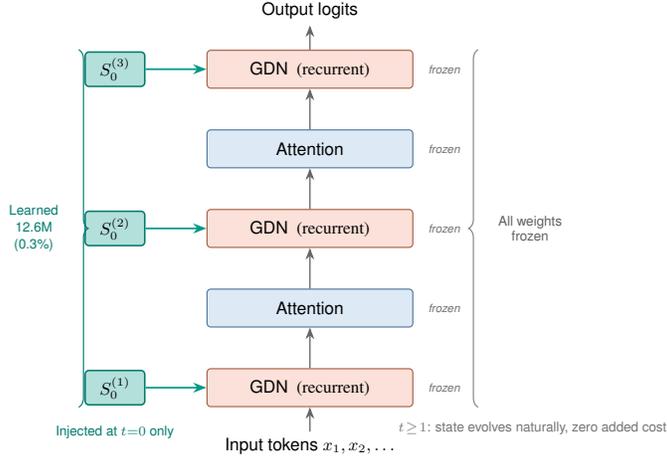
\begin{figure}[t]
  \centering
  \definecolor{s0color}{HTML}{009988}
  \definecolor{frozencolor}{HTML}{BBBBBB}
  \definecolor{attncolor}{HTML}{77AADD}
  \definecolor{gdncolor}{HTML}{EE8866}
  \definecolor{arrowcolor}{HTML}{009988}
  \scalebox{0.72}{%
  \begin{tikzpicture}[
      >=Stealth,
      layer/.style={
        minimum width=3.8cm, minimum height=0.7cm,
        rounded corners=2pt, draw=#1!70!black, line width=0.6pt,
        fill=#1!25, font=\small\sffamily},
      s0box/.style={
        minimum width=1.1cm, minimum height=0.5cm,
        rounded corners=2pt, draw=s0color!80!black, line width=0.8pt,
        fill=s0color!30, font=\small\bfseries\sffamily},
      froztag/.style={font=\tiny\sffamily\color{frozencolor!60!black}},
      dataarrow/.style={->, line width=0.7pt, color=black!60},
      s0arrow/.style={->, line width=0.9pt, color=arrowcolor},
    ]
    \def\layersep{0.75}
    \node[font=\small\sffamily] (input) at (0, 0) {Input tokens $x_1, x_2, \ldots$};
    \node[layer=gdncolor, above=0.45cm of input] (gdn1) {GDN\enspace\textnormal{(recurrent)}};
    \node[froztag, right=0.15cm of gdn1.east] (fw1) {\textit{frozen}};
    \node[layer=attncolor, above=\layersep cm of gdn1] (attn1) {Attention};
    \node[froztag, right=0.15cm of attn1.east] (fa1) {\textit{frozen}};
    \node[layer=gdncolor, above=\layersep cm of attn1] (gdn2) {GDN\enspace\textnormal{(recurrent)}};
    \node[froztag, right=0.15cm of gdn2.east] (fw2) {\textit{frozen}};
    \node[layer=attncolor, above=\layersep cm of gdn2] (attn2) {Attention};
    \node[froztag, right=0.15cm of attn2.east] (fa2) {\textit{frozen}};
    \node[layer=gdncolor, above=\layersep cm of attn2] (gdn3) {GDN\enspace\textnormal{(recurrent)}};
    \node[froztag, right=0.15cm of gdn3.east] (fw3) {\textit{frozen}};
    \node[font=\small\sffamily, above=0.45cm of gdn3] (output) {Output logits};
    \draw[dataarrow] (input) -- (gdn1);
    \draw[dataarrow] (gdn1) -- (attn1);
    \draw[dataarrow] (attn1) -- (gdn2);
    \draw[dataarrow] (gdn2) -- (attn2);
    \draw[dataarrow] (attn2) -- (gdn3);
    \draw[dataarrow] (gdn3) -- (output);
    \node[s0box] (s01) at (-3.6, 0.45cm |- gdn1) {$\svec^{(1)}$};
    \node[s0box] (s02) at (-3.6, 0cm |- gdn2) {$\svec^{(2)}$};
    \node[s0box] (s03) at (-3.6, 0cm |- gdn3) {$\svec^{(3)}$};
    \draw[s0arrow] (s01.east) -- (gdn1.west);
    \draw[s0arrow] (s02.east) -- (gdn2.west);
    \draw[s0arrow] (s03.east) -- (gdn3.west);
    \draw[decorate, decoration={brace, amplitude=5pt, mirror},
          s0color!80!black, line width=0.6pt]
      ([xshift=-3pt]s01.south west |- gdn1.south) -- ([xshift=-3pt]s03.north west |- gdn3.north)
      node[midway, left=7pt, align=center, font=\scriptsize\sffamily,
           text=s0color!80!black] {Learned\\[1pt]12.6M\\[1pt](0.3\%)};
    \draw[decorate, decoration={brace, amplitude=5pt},
          frozencolor!70!black, line width=0.6pt]
      ([xshift=6pt]fw1.south east |- gdn1.south) --
      ([xshift=6pt]fw3.north east |- gdn3.north)
      node[midway, right=7pt, align=center, font=\scriptsize\sffamily,
           text=frozencolor!50!black] {All weights\\frozen};
    \node[font=\scriptsize\sffamily, text=s0color!80!black,
          below=0.25cm of s01, align=center]
      {Injected at $t{=}0$ only};
    \node[font=\scriptsize\sffamily, text=black!55,
          below right=0.15cm and -0.4cm of gdn1.south east, align=left]
      {$t \!\ge\! 1$: state evolves naturally, zero added cost};
  \end{tikzpicture}%
  }
  \caption{%
    \textbf{Computation graph for \method.}
    The learned initial state $\svec$ (teal) is injected into each recurrent layer
    before the first token. After $t{=}1$, it is absorbed into the running state and
    adds zero computational overhead. All model weights remain frozen.
  }
  \label{fig:method}
\end{figure}

Figure~\ref{fig:method} illustrates the computation graph.

\section{Experimental Setup}
\label{sec:setup}

\subsection{Models}
\label{sec:setup_models}

We evaluate on two hybrid architectures.
\textbf{Qwen3.5-4B}~\cite{qwen3_5,gated_deltanet} interleaves 24 \gdn layers with 8
attention layers (${\sim}$3:1 ratio); each GDN layer carries a state matrix
$\state \in \mathbb{R}^{32 \times 128 \times 128}$, totaling 12.6M state
parameters (0.3\% of the model).
\textbf{\falcon-7B}~\cite{falconh1} is a Mamba-2 hybrid where every layer
processes tokens through both attention and Mamba-2 heads in parallel, totaling
34.6M state parameters (0.5\%).
We also evaluate scaling across Qwen3.5-\{0.8B, 2B, 9B\} with identical
hyperparameters. All models run in \texttt{no\_thinking} mode (chain-of-thought generation disabled).

\subsection{Benchmarks}
\label{sec:setup_benchmarks}

Our primary benchmark is HumanEval~\cite{humaneval} (164 problems). We reserve
problems 0--79 for training data collection and evaluate on the held-out set of
problems 80--163 ($n{=}84$). Training data consists of execution-verified
correct solutions sampled from the frozen base model. The main HumanEval
pipeline keeps at most one passing completion per train problem, yielding
roughly 48 verified solutions total across the 80 training problems. For
cross-domain evaluation we test on MATH-500~\cite{math500},
GSM8K~\cite{gsm8k}, and Spider~\cite{spider} (text-to-SQL) using standard splits.

\subsection{Baselines}
\label{sec:setup_baselines}

We compare against LoRA~\cite{lora} as the primary baseline. On Qwen3.5-4B,
LoRA targets $\{$q,k,v,o$\}$\_proj of the 8 attention layers (rank~24, 4.7M
parameters). On \falcon-7B, LoRA targets $\{$q,k,v,in$\}$\_proj across all
layers (rank~24, 22.2M parameters); \texttt{in\_proj} belongs to the Mamba-2
mixer, so \falcon LoRA adapts \emph{both} attention and recurrent pathways. We
select the best of 4 configurations (2 learning rates $\times$ 2 ranks).
As a negative control, we apply prefix tuning with matched parameter count to
Qwen2.5-3B-Instruct~\cite{qwen2_5}, a pure Transformer (30 virtual tokens,
\texttt{prefix\_projection=True}, 3 learning rates $\times$ 3 seeds).
Full configurations are in Appendix~\ref{app:hyperparams}.

\paragraph{Comparison caveats.}
The primary Qwen comparison uses \methodshort at 12.6M versus LoRA rank-24 at
4.7M; Section~\ref{sec:results_matched} tests matched budgets (rank-64 LoRA
degrades by $-15.5$~\pp{} in the same setup). Unless otherwise noted, Qwen
LoRA numbers in the main paper refer to this best rank-24 baseline, while
higher-rank LoRA appears only in Section~\ref{sec:results_matched}.
\methodshort also requires execution-verified solutions, a stronger data
assumption than LoRA, and the Falcon comparison remains a supportive 3-seed
result rather than a definitive superiority claim.

\subsection{Statistical Protocol}
\label{sec:setup_stats}

All training hyperparameters are in Appendix~\ref{app:hyperparams}.
Primary comparisons use 10 random seeds. We report mean $\pm$ std and
test significance with Welch's $t$-test (unequal variances). Unless otherwise
stated, hypothesis tests in the main paper are two-sided; paired $t$-tests are
used only for the dedicated \methodshort versus state-offset rerun.
Comparisons with only 3 seeds (Falcon, 2B, 9B, and pass@$k$) are reported as
supportive evidence rather than headline superiority claims. GPU
floating-point non-determinism introduces ${\sim}$2~\pp{} baseline variation
across seeds; this is absorbed into the reported standard deviations.

\paragraph{Claim hierarchy.}
The paper has one anchor claim: on Qwen3.5-4B HumanEval, \methodshort beats the
main LoRA baseline over 10 seeds. Falcon tests architectural transfer,
MATH-500 and GSM8K test domain transfer, Spider marks a failure case, and the
matched-budget LoRA plus Transformer prefix-tuning controls test alternative
explanations. We use the strongest statistical language only for the
comparisons that carry the central claim.

\section{Results}
\label{sec:results}

\subsection{Main Comparison}
\label{sec:results_main}

Table~\ref{tab:main} summarizes the main empirical result. On Qwen3.5-4B,
\methodshort improves greedy HumanEval pass@1 from 48.8\% to 72.2\%, a gain of
$\mainresult \pm 1.7$~\pp{} over 10 seeds.\footnote{Deltas are computed as the mean of per-seed improvements, which may differ slightly from the difference of rounded group means.} Against the strongest Qwen LoRA
baseline (rank~24),
this yields a $\loragap$~\pp{} advantage ($p = 0.000056$) with roughly
$3\times$ lower variance (1.7 vs.\ 5.1~\pp). State-offset attains the highest
absolute score on Qwen3.5, but does so by paying a per-step inference cost;
\methodshort remains the best zero-inference-overhead operating point.

The same table also shows that the result is not specific to one hybrid family.
On \falcon{}-7B, \methodshort reaches $71.8\% \pm 1.3$ with architecture-specific
$\alpha{=}0.65$; LoRA reaches $71.4\% \pm 2.4$. The two methods
are statistically indistinguishable at this sample size (3 seeds) while
\methodshort avoids weight merging and again exhibits lower variance. We treat
this as evidence of competitiveness on Falcon rather than a superiority claim.

In a separate 3-seed sampled decoding evaluation, \methodshort reaches pass@10
of 88.5\% versus 66.7\% for LoRA (Table~\ref{tab:passk} in the appendix), a
gap that grows with $k$. LoRA's pass@5 is tied with the untrained baseline.

\begin{table}[t]
\centering
\caption{HumanEval main results on held-out problems 80--163. Qwen results use 10 seeds; Falcon results use 3 seeds unless noted otherwise.}
\label{tab:main}
\small
\resizebox{\textwidth}{!}{%
\begin{tabular}{@{}lcccl@{}}
\toprule
\textbf{Method} & \textbf{Params\textsuperscript{$\dagger$}} & \textbf{Qwen3.5-4B (\gdn)} & \textbf{\falcon{}-7B (Mamba-2)} & \textbf{Overhead} \\
\midrule
Baseline           & ---   & 48.8\%                         & 40.5\%                          & ---      \\
LoRA (best config) & 4.7M / 22.2M  & 61.5\% $\pm$ 5.1\%            & 71.4\% $\pm$ 2.4\%             & Merge    \\
State-offset       & 12.6M / ---   & 74.8\% $\pm$ 3.6\%            & ---                             & Per-step \\
\methodshort (ours) & 12.6M / 34.6M & \textbf{72.2\% $\pm$ 1.7\%} & \textbf{71.8\% $\pm$ 1.3\%} & None \\
\bottomrule
\end{tabular}%
}
\par\smallskip\noindent{\footnotesize $\dagger$~Params shown as Qwen3.5 / \falcon.
\falcon has 44 Mamba-2 layers (vs.\ 24 GDN layers in Qwen3.5), yielding more state parameters (34.6M vs.\ 12.6M) and more LoRA targets (22.2M vs.\ 4.7M).
The state-offset accuracy comes from the paired 10-seed run in Table~\ref{tab:offset}: 47.6\% baseline + 27.1~\pp = 74.7\%.}
\end{table}

\subsection{Parameter-Matched Comparison}
\label{sec:results_matched}

The primary LoRA baseline uses rank~24 (4.7M parameters), roughly one-third of
\methodshort's 12.6M. A natural objection is that \methodshort wins because it
has more capacity. We test this directly by training LoRA at rank~48 (9.4M)
and rank~64 (12.6M, exactly matching \methodshort), using the same data,
learning rate, and training steps (Table~\ref{tab:matched}).

Increasing LoRA rank makes things worse, not better. Rank~48 averages
$+2.1 \pm 17.0$~\pp{} across 10 seeds, with one catastrophic seed collapsing
to 4.8\% accuracy. Rank~64, at matched parameter budget, degrades by
$-15.5 \pm 18.9$~\pp: 8 of 10 seeds are negative. The pattern is consistent
with overfitting in this small-data regime.
At matched budget, the gap between \methodshort ($+23.6$~\pp) and LoRA rank-64 ($-15.5$~\pp) is 39~\pp: identical parameter counts organized as recurrent states versus weight matrices produce opposite outcomes in this setting.

\begin{table}[t]
\centering
\caption{Parameter-matched LoRA comparison on Qwen3.5-4B HumanEval (10 seeds each).}
\label{tab:matched}
\small
\begin{tabular}{@{}lrccc@{}}
\toprule
\textbf{Method} & \textbf{Params} & \textbf{Mean $\Delta$} & \textbf{Std} & \textbf{Seeds negative} \\
\midrule
LoRA r24         & 4.7M  & $+12.7$~\pp & 5.1~\pp  & 0/10 \\
LoRA r48         & 9.4M  & $+2.1$~\pp  & 17.0~\pp & 1/10 collapsed \\
LoRA r64         & 12.6M & $-15.5$~\pp & 18.9~\pp & 8/10 \\
\methodshort & 12.6M & $\mathbf{+23.6}$~\pp & 1.7~\pp & 0/10 \\
\bottomrule
\end{tabular}
\end{table}

\subsection{Cross-Architecture Transfer}
\label{sec:results_crossarch}

Qwen3.5 (interleaved \gdn + attention) and \falcon{}-7B (parallel Mamba-2 +
attention) differ in both recurrence family and hybrid topology, making the
cross-architecture result a stronger test. The Falcon result in
Table~\ref{tab:main} shows \methodshort remains competitive even when LoRA
targets the Mamba-2 \texttt{in\_proj} pathway, but we regard this as
supportive evidence until a larger-seed replication is run.

\subsection{Scaling Behavior}
\label{sec:results_scaling}

\begin{figure}[t]
  \centering
  \includegraphics[width=\textwidth]{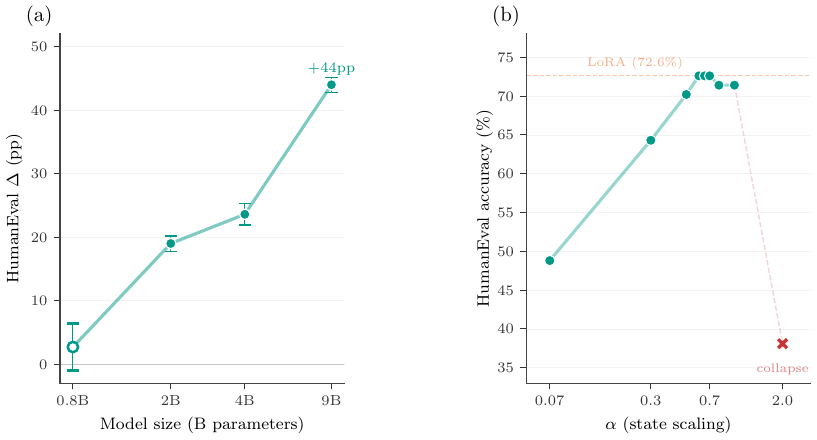}
  \caption{%
    \textbf{Scaling and architecture-specific tuning.}
    (a)~Performance gains increase monotonically with model scale, from
    $+2.6$~\pp{} at 0.8B to $+44.0$~\pp{} at 9B.
    Error bars show $\pm 1$ standard deviation across seeds.
    The 9B model achieves 76.1\% absolute accuracy from a 32.1\% baseline,
    suggesting larger models have more latent capability to unlock via
    state initialization.
    (b)~\falcon alpha sweep: the default $\alpha{=}0.07$ yields only $+8.3$~\pp,
    but architecture-specific tuning to $\alpha{=}0.6$--$0.7$ reaches $71.8\%$,
    matching LoRA's $71.4\%$. Large $\alpha$ ($\ge 2.0$) collapses performance.
  }
  \label{fig:scaling}
\end{figure}

Figure~\ref{fig:scaling}(a) shows a strong scale trend within the Qwen3.5 family.
The 0.8B model exhibits only a directional gain ($+2.6 \pm 3.7$~\pp,
$p{=}0.076$), but performance improves sharply at 2B
($+19.0 \pm 1.2$~\pp, $p{=}0.001$), 4B ($+23.6 \pm 1.7$~\pp), and 9B
($+44.0 \pm 1.2$~\pp, $p{=}0.0002$).
Gains increase with model scale, from $+2.6$~\pp{} at 0.8B to $+44.0$~\pp{} at 9B, though the non-monotonic baseline complicates scaling analysis. The 9B baseline (32.1\%) is lower than the 4B (48.8\%) because suppressing chain-of-thought degrades the larger model more; $\svec$ recovers it to 76.1\%. Full scaling data are in Appendix~\ref{app:probing}.

\subsection{State-Offset Comparison}
\label{sec:results_offset}

A per-step state-offset variant~\cite{state_offset} achieves $+27.1$~\pp{}
(Table~\ref{tab:offset}), the strongest absolute accuracy, but at per-step
inference cost. Table~\ref{tab:offset} uses a dedicated paired rerun for the
two state-based methods, which is why \methodshort appears there as
$+23.5 \pm 4.0$~\pp{} rather than the main-run $+23.6 \pm 1.7$~\pp{}. In that
paired rerun, \methodshort captures 87\% of the offset's gain ($+23.5$ of
$+27.1$~\pp) at zero runtime cost. The paired rerun is only for comparing
\methodshort with state-offset; LoRA is shown in Table~\ref{tab:offset} as the
main rank-24 reference baseline from Table~\ref{tab:main}.

\begin{table}[t]
\centering
\caption{State-based comparison on Qwen3.5-4B HumanEval. \methodshort and state-offset use a dedicated paired 10-seed rerun; the LoRA row is included only as the main rank-24 reference baseline from Table~\ref{tab:main}.}
\label{tab:offset}
\small
\begin{tabular}{@{}lcccc@{}}
\toprule
\textbf{Method} & \textbf{Mean $\Delta$} & \textbf{Std} & \textbf{Gap vs.\ main LoRA} & \textbf{Overhead} \\
\midrule
LoRA (best)              & $+12.7$~\pp & 5.1~\pp & ---          & Merge    \\
\methodshort             & $+23.5$~\pp & 4.0~\pp & $+10.8$~\pp & None     \\
State-offset             & $+27.1$~\pp & 3.6~\pp & $+14.4$~\pp & Per-step \\
\bottomrule
\end{tabular}
\end{table}

\subsection{Cross-Domain Transfer}
\label{sec:results_crossdomain}

Cross-domain gains are smaller but statistically significant: MATH-500
improves by $+4.8 \pm 1.4$~\pp{} (two-sided $p{=}0.00002$, 8~seeds) and GSM8K
by $+2.8 \pm 1.6$~\pp{} (two-sided $p{=}0.0003$, 10~seeds). Spider text-to-SQL
shows no transfer ($+0.0$~\pp{} across five alpha values in a single-seed
alpha sweep), consistent with the trajectory-steering mechanism: SQL queries
have low early-token diversity, so the initial-state perturbation has little
to steer. We treat Spider as a boundary-condition observation rather than a
formal significance result.

\subsection{Architecture-Specific Alpha}
\label{sec:results_alpha}

The alpha sweep (Appendix~\ref{app:alpha}) explains why the first Falcon result understated the
method's potential. The Qwen-optimal default $\alpha{=}0.07$ improves FalconH1
by only $+8.3$~\pp{}, but performance rises steeply as the scale increases,
plateauing at $0.60$--$0.70$ at $71.8\%$ accuracy, matching LoRA's $71.4\%$. Pushing $\alpha$ too far
eventually collapses the model: at $\alpha{=}2.0$, accuracy falls below the
baseline.

Full alpha sweep values are in Appendix~\ref{app:alpha}.

\subsection{Transformer Negative Control}
\label{sec:results_transformer}

Prefix tuning on a pure Transformer (Qwen2.5-3B) degrades performance by $-13.9$~\pp{} (all 9 configurations negative, 95\% CI upper bound $-5.3$\%); the gap to hybrid \methodshort is $+37.5$~\pp, consistent with recurrence playing a key role.

\section{Mechanistic Analysis}
\label{sec:mechanistic}

\method produces large accuracy gains on two hybrid architectures while
matched-parameter prefix tuning on a pure Transformer fails entirely
($-13.9$~\pp). \emph{Why} does tuning the recurrent initial state succeed
where prefix tuning does not? We examine persistence analysis, first-character divergence, linear probing,
and architecture-specific gating dynamics.

\subsection{The Launch Vector Effect}
\label{sec:mech_launch}

Consider the \gdn update (Eq.~\ref{eq:gdn_update}) applied over $T$ prompt
tokens. $\svec$'s contribution to the state at position $T$ is
\begin{equation}
  S_T^{(\svec)} \;=\; \svec \prod_{t=1}^{T} G_t,
  \qquad G_t = \alpha_t (I - \beta_t\, k_t k_t^\top),
  \label{eq:decay}
\end{equation}
where each $G_t$ is the per-step gating matrix. When the decay gates
$\alpha_t$ are close to 1 and the write strengths $\beta_t$ are small, the
product $\prod G_t$ retains a nonzero component of $\svec$; when gating is
aggressive, $\svec$ decays toward zero. This yields two testable predictions: the direct KL influence of $\svec$ on output logits should
decay roughly exponentially through the prompt, at a rate set by the average
gate magnitude, and architectures with different gating statistics
(GDN vs.\ Mamba-2) should require proportionally different alpha scaling to
achieve the same effective perturbation, explaining the 10$\times$ gap
between optimal alpha values. We test both below.

\subsection{Persistence Analysis}
\label{sec:mech_persistence}

Equation~\ref{eq:decay} predicts exponential decay of $\svec$'s direct
influence. We test this by comparing forward passes with the tuned state
against forward passes with the default zero state, both conditioned on the
same prompt tokens. We measure the KL divergence ratio between the two output
distributions at each position. The influence decays through the prompt
context, reaching 0.03\% by the final prompt token, matching the
predicted exponential profile. By the time generation begins, the tuned state
and zero state produce almost identical next-token distributions \emph{over
the prompt}. The signal has not vanished. It has been absorbed into the
recurrent state as a low-magnitude directional bias: a slight but consistent
skew in the hidden representation rather than a large perturbation to the
output logits.

\begin{figure}[t]
  \centering
  \includegraphics[width=\textwidth]{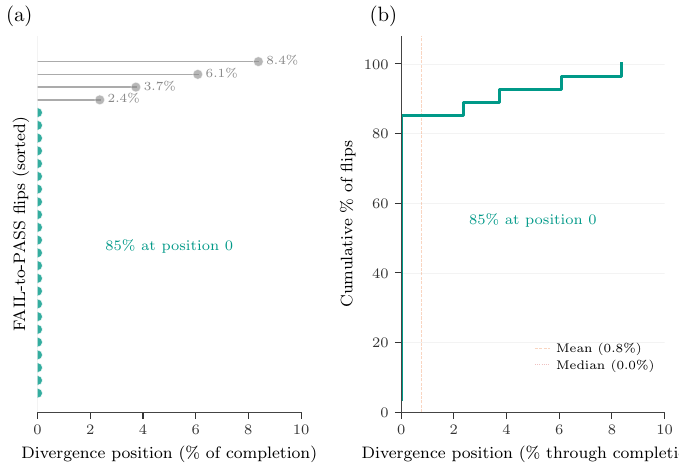}
  \caption{%
    \textbf{First-character divergence in FAIL-to-PASS flips} (27 flips,
    single-seed Qwen3.5-4B).
    (a)~23 of 27 corrected solutions (85\%) diverge from baseline at
    character position~0, the very first generated character.
    (b)~Cumulative distribution of divergence positions: all 27 flips diverge
    within the first 10\% of the completion (mean 0.76\%, median 0.0\%).
    $\svec$ shifts the output distribution at the first opportunity;
    autoregressive decoding amplifies this into a qualitatively different
    solution.
  }
  \label{fig:mechanistic}
\end{figure}

\subsection{First-Character Divergence}
\label{sec:mech_divergence}

If $\svec$ acts as a soft prompt in recurrent memory, its effect should be
visible the moment the model begins generating. We test this by examining all
27 FAIL-to-PASS flips from a single-seed Qwen3.5-4B evaluation and recording the
first character position at which the tuned and baseline greedy outputs differ. Of these 27 flips, 23 (85\%) diverge at character position~0, the very first generated character. The remaining 4 diverge at positions 16, 28, 32, and 36; all 27 diverge within the first 10\% of the completion. A sign test for concentration within the first 10\% yields $p < 10^{-8}$. $\svec$ does not gradually steer generation over many tokens. It shifts the model's output distribution at the first opportunity, and autoregressive decoding amplifies that initial divergence into a qualitatively different solution. This is exactly what the persistence analysis predicts: by the time generation begins,
$\svec$ has been compressed into the recurrent state as a small directional
bias that is just large enough to flip the argmax at the first generated character.

\subsection{Architecture-Specific Gating and Probing}
\label{sec:mech_probing}

Equation~\ref{eq:decay} also predicts that architectures with different gating
statistics can require different alpha scaling. GDN layers in Qwen3.5 use both
a scalar decay $\alpha_t$ and a key-dependent erasure term
$\beta_t k_t k_t^\top$, whereas Mamba-2 layers in \falcon use scalar gating
through SSD. The observed 10$\times$ gap (0.07 for Qwen3.5 vs.\ 0.65 for
\falcon; Appendix~\ref{app:alpha}) shows that the same initial-state scale
does not transfer unchanged across recurrence families. A linear probe trained to predict solution
correctness from intermediate representations (PCA-reduced to 64 dimensions;
see Appendix~\ref{app:probing} for full protocol) achieves
$\AUC = 0.93$ when reading from the recurrent state versus $\AUC = 0.90$ from
the residual stream. This 2.5-point gap (0.930 vs.\ 0.905) is descriptive
rather than inferential, but it suggests that the recurrent state encodes
task-relevant information at least as directly as the residual stream.

\section{Related Work}
\label{sec:related}

\subsection{PEFT for State Space Models}
\label{sec:rel_peft_ssm}

The PEFT literature for SSMs shares a common finding: on diagonal-state architectures, LoRA wins.
\citet{galim_peft_ssm} showed this most directly at ICML~2025, comparing LoRA against initial-state tuning on Mamba-1 (their appendix extends to Jamba and Mamba-II with scalar or small-dimensional states).
MambaPEFT~\cite{mamba_peft} evaluated 20 PEFT variants without testing initial-state tuning at all.
ProDiaL~\cite{prodial} and SSMLoRA~\cite{ssmlora} explored projector and hybrid adaptations; Memba~\cite{memba} introduced membrane gating.
None of these works tested matrix-valued states.
Prefix tuning~\cite{prefix_tuning} prepends learnable key-value vectors to Transformer attention; we use it as a negative control (Section~\ref{sec:results_transformer}).

On \gdn and Mamba-2, where states are full matrices (${\sim}$524K entries per layer), we observe the opposite: \method outperforms LoRA by \loragap~\pp{} ($p < 0.001$) on Qwen3.5.
The reversal supports a state-expressiveness hypothesis: a 524K-entry matrix can encode enough task-relevant structure for PEFT, while a diagonal state does not appear to.

\subsection{State-Based Adaptation}
\label{sec:rel_state}

State-offset tuning~\cite{state_offset} adds a per-step learned offset to Mamba-1 hidden states (up to 1.4B parameters, NLU benchmarks).
\citet{rwkv_state_tuning} tune RWKV-7 states with kernel upscaling and decorrelated backpropagation, framed as test-time scaling rather than PEFT; we adopt the name \methodshort tuning to distinguish our gradient-based approach from their broader ``State Tuning'' framework.
Concurrently, \citet{midi_rwkv} compare state tuning against LoRA on MIDI-RWKV for music infilling, finding state tuning competitive at low ranks but with higher variance than LoRA across seeds.
Lina-Speech~\cite{lina_speech} optimizes $\svec$ of Gated Linear Attention layers for TTS voice cloning, the closest prior art to our mechanism but in a different domain and without LoRA comparison.

Independent work establishes that recurrent states \emph{encode} task identity even without optimization.
State Soup~\cite{state_soup} showed that in-context-learned states in Mamba-2.8B cluster by task and that interpolating them improves few-shot performance.
We go further: instead of interpolating states that arise from in-context learning, we optimize them directly via gradient descent.

We extend the offset idea to \gdn, where our offset variant reaches $+27.1$~\pp{} (beating LoRA at $p < 0.001$); the pure-$\svec$ variant trades peak accuracy for zero inference overhead.
No prior work validates state-based PEFT across multiple recurrence families in a single study.

\subsection{Hybrid Architectures}
\label{sec:rel_hybrid}

Mamba-2~\cite{mamba2} introduced structured state space duality with matrix-valued states; \gdn~\cite{gated_deltanet} applied the delta rule to linear attention.
Qwen3.5 interleaves \gdn layers with standard attention at a roughly 3:1 ratio; \falcon~\cite{falconh1} places Mamba-2 and attention heads in parallel within every layer.
Mamba-3~\cite{mamba3} and \citet{hybrid_analysis} study how to design and analyze such hybrids.
The recurrent state ($\state \in \stateshape$, updated at every token) is the adaptation surface our method targets; \method tunes it while leaving all weights frozen.

\section{Discussion and Limitations}
\label{sec:discussion}

\paragraph{Code--math gap.}
\method is strongest on code generation ($\mainresult$~\pp{} on HumanEval).
Cross-domain transfer to MATH-500 ($+4.8$~\pp, two-sided $p{=}0.00002$, 8
seeds) and GSM8K ($+2.8$~\pp, two-sided $p{=}0.0003$, 10 seeds) is
statistically significant but substantially smaller. Spider text-to-SQL
($+0.0$~\pp) shows no transfer to structured-output tasks.

\paragraph{Diagonal vs.\ matrix states.}
Our results are on architectures with matrix-valued states (\gdn, Mamba-2). \citet{galim_peft_ssm} showed that initial-state tuning underperforms LoRA on diagonal-state Mamba-1, consistent with a state-expressiveness threshold below which initial-state tuning cannot compete with weight adaptation.

\paragraph{Training data.}
\method requires execution-verified correct solutions (${\sim}$48 verified
completions for our HumanEval setup). That is a small training set relative to
the breadth of our claims, so the headline result should be read as small-data
evidence that recurrent state is a strong adaptation surface in this regime.
Applying \methodshort where verification is expensive remains untested. LoRA
imposes no such constraint. HumanEval problems may appear in pretraining data;
cross-domain gains on MATH-500 and GSM8K argue against a memorization-only
explanation, and the null result on Spider further bounds the scope of
transfer.
Training takes 3 minutes on a single A10G GPU; as few as 25 solutions suffice for stable gains (Appendix~\ref{app:data_efficiency}).

\paragraph{Evidence strength and compute budget.}
The core result is overdetermined on Qwen. The thinner comparisons serve a
different job: Falcon checks architectural transfer, Spider marks a failure
mode, and the auxiliary sweeps show how far the effect moves beyond the anchor
setting. Falcon still uses 3 seeds, Spider is a single-seed alpha sweep, and we
did not keep broadening the benchmark grid once the project hit a fixed
personal GPU budget rather than institutional cluster access. We therefore
present these experiments as transfer checks and boundary conditions, not as a
complete benchmark sweep over every competing method and category.

\section{Conclusion}
\label{sec:conclusion}

\method optimizes only the initial recurrent state of hybrid language models,
adding zero inference overhead. On Qwen3.5-4B, \methodshort outperforms LoRA by
$\loragap$~\pp{} ($p < 0.001$); on \falcon-7B it is statistically
indistinguishable from LoRA in a 3-seed comparison ($71.8\%$ vs.\ $71.4\%$). A
prefix-tuning control on a pure Transformer fails under all configurations.
Cross-domain transfer to MATH-500 and GSM8K is significant ($p < 0.001$),
though the method does not transfer to structured-output SQL.
Matrix-valued recurrent states~\cite{galim_peft_ssm} appear to provide an
expressive adaptation surface that LoRA leaves unused in this setting;
\methodshort is a competitive zero-inference-overhead PEFT method for
exploiting it.

\bibliographystyle{plainnat}
\bibliography{references}

\appendix

\section{Hyperparameters}
\label{app:hyperparams}

Table~\ref{tab:hyperparams} lists all hyperparameters for the \methodshort and LoRA experiments reported in the main text.

\begin{table}[h!]
\centering
\caption{Hyperparameters for the Qwen3.5 and FalconH1 experiments.}
\label{tab:hyperparams}
\footnotesize
\begin{tabular}{@{}lcc@{}}
\toprule
\textbf{Hyperparameter} & \textbf{\methodshort} & \textbf{LoRA} \\
\midrule
Learning rate     & $1 \times 10^{-3}$    & $5 \times 10^{-4}$ (Qwen) / $1 \times 10^{-4}$ (Falcon) \\
Optimizer         & Adam                   & Adam \\
Training steps    & 20                     & 50 \\
Batch size        & 1                      & 1 \\
L2 regularization & $5 \times 10^{-4}$    & --- \\
Alpha scaling     & 0.07 (GDN) / 0.65 (Mamba-2) & --- \\
LoRA rank         & ---                    & 24 (primary) / 48, 64 (matched-budget) \\
LoRA target       & ---                    & \{q,k,v,o\}\_proj (Qwen) / \{q,k,v,in\}\_proj (Falcon)\textsuperscript{$\ddagger$} \\
Loss              & Completion-only        & Completion-only \\
Training data     & Correct HumanEval train solutions & Correct HumanEval train solutions \\
Precision         & bf16                   & bf16 \\
Hardware          & NVIDIA A10G (24\,GB)   & NVIDIA A10G (24\,GB) \\
Training time     & ${\sim}$3 min          & ${\sim}$5 min \\
Trainable params  & 12.6M (Qwen) / 34.6M (Falcon)\textsuperscript{$\dagger$} & 4.7M (Qwen) / 22.2M (Falcon) \\
\bottomrule
\end{tabular}
\par\smallskip\noindent{\footnotesize $\ddagger$~On \falcon-7B, \texttt{in\_proj} belongs to the Mamba-2 mixer, so the Falcon LoRA baseline adapts both attention and recurrent pathways.}
\par\noindent{\footnotesize $\dagger$~On \falcon-7B, \methodshort tunes 34.6M parameters (0.5\%) due to larger Mamba-2 state matrices.}
\end{table}

\section{Layer Sensitivity Ablation}
\label{app:layers}

Table~\ref{tab:layers} breaks down the contribution of each layer group.
The ordering early $>$ middle $>$ late is consistent, with gaps of approximately 5~\pp between adjacent groups (${\sim}3\sigma$ separation).
Early layers (0--7) alone recover $+23.8$~\pp, matching the full-model $+22.6$~\pp at one-third the parameter count (4.2M vs.\ 12.6M).
Late layers yield the smallest gain ($+13.1$~\pp) but produce zero degradations, making them the safest choice under a strict ``do no harm'' constraint.
If parameter budget is the binding concern, tuning only the first 8 layers is a practical default.

\begin{table}[h!]
\centering
\caption{Layer sensitivity ablation on Qwen3.5-4B HumanEval (single seed).}
\label{tab:layers}
\small
\begin{tabular}{@{}lrcrc@{}}
\toprule
\textbf{Layers Tuned} & \textbf{\# Layers} & \textbf{Greedy $\Delta$} & \textbf{Degraded} & \textbf{Params} \\
\midrule
Early (0--7)   & 8  & $+23.8$~\pp & 2 & 4.2M \\
Middle (8--15)  & 8  & $+19.0$~\pp & 3 & 4.2M \\
Late (16--23)   & 8  & $+13.1$~\pp & 0 & 4.2M \\
All (0--23)     & 24 & $+22.6$~\pp & 4 & 12.6M \\
\bottomrule
\end{tabular}
\end{table}

\section{Per-Seed Results}
\label{app:perseed}

Tables~\ref{tab:lora_perseed} and~\ref{tab:falcon_perseed} report per-seed results for full reproducibility.

\begin{table}[h!]
\centering
\caption{Per-seed LoRA results on Qwen3.5-4B HumanEval (best configuration: lr $= 5{\times}10^{-4}$, rank 24, 50 steps).}
\label{tab:lora_perseed}
\small
\begin{tabular}{@{}rrl@{}}
\toprule
\textbf{Seed} & \textbf{LoRA $\Delta$} & \textbf{Degraded} \\
\midrule
42   & $+9.5$~\pp  & 0  \\
7    & $+17.9$~\pp & 0  \\
123  & $+20.2$~\pp & 0  \\
1337 & $+15.5$~\pp & 10 \\
999  & $+17.9$~\pp & 6  \\
314  & $+11.9$~\pp & 11 \\
256  & $+8.3$~\pp  & 12 \\
777  & $+3.6$~\pp  & 14 \\
555  & $+10.7$~\pp & 7  \\
2024 & $+11.9$~\pp & 7  \\
\midrule
\textbf{Mean} & $+12.7 \pm 5.1$~\pp & --- \\
\bottomrule
\end{tabular}
\end{table}

\begin{table}[h!]
\centering
\caption{\falcon-7B per-seed results at $\alpha{=}0.65$ on HumanEval held-out problems 80--163.}
\label{tab:falcon_perseed}
\small
\begin{tabular}{@{}rcccc@{}}
\toprule
\textbf{Seed} & \textbf{Baseline} & \textbf{Tuned} & \textbf{$\Delta$} & \textbf{Degraded} \\
\midrule
42  & 40.5\% & 72.6\% & $+32.1$~\pp & --- \\
7   & 40.5\% & 70.2\% & $+29.8$~\pp & --- \\
123 & 40.5\% & 72.6\% & $+32.1$~\pp & --- \\
\midrule
\textbf{Mean} & 40.5\% & 71.8\% & $+31.3 \pm 1.3$~\pp & --- \\
\bottomrule
\end{tabular}
\end{table}

\section{Training Data Efficiency}
\label{app:data_efficiency}

We vary the number of correct training solutions used to optimize $\svec$ (3 seeds each, Qwen3.5-4B, HumanEval).
With only 10 solutions, the method already produces a $+25.4 \pm 6.8$~\pp gain, though variance is high.
At 25 solutions the mean stabilizes to $+22.6 \pm 3.2$~\pp, and doubling to 50 solutions yields the same mean ($+22.6$~\pp) with no further benefit.
The method is data-efficient: 25 correct solutions suffice for stable gains, and even 10 are enough to see large (if noisy) improvements.
Practitioners with limited labeled data can expect usable results from a small pool of verified solutions.

\section{Alpha Sweep}
\label{app:alpha}

On Qwen3.5-4B (\gdn), we sweep the alpha scaling factor using a sampled metric (temperature 0.7, single seed).
Performance plateaus across a wide range: $\alpha{=}0.03$ through $\alpha{=}0.15$ all yield $+30$--$+34$~\pp sampled gains, with the peak at $\alpha{=}0.07$ ($+33.9$~\pp sampled).
These sampled numbers are higher than the greedy 10-seed mean of $\mainresult$~\pp reported in the main text because sampling at temperature 0.7 amplifies the effect of the learned prior.
The broad plateau means the method is not sensitive to alpha within this range on \gdn.

The \falcon-7B sweep tells a different story.
Applying the \gdn-optimal $\alpha{=}0.07$ to Mamba-2 yields only $+8.3$~\pp, while the optimal range of $\alpha{=}0.60$--$0.70$ reaches $71.8\%$ accuracy, a $4\times$ larger gain.
The roughly $10\times$ scale difference between the two architectures ($0.07$ vs.\ $0.65$) shows that the effective state scale is architecture-specific. \gdn combines scalar decay with key-dependent erasure, while Mamba-2 uses SSD gating; our experiments do not isolate which part of the update drives the alpha shift, only that transferring the Qwen setting unchanged leaves a large amount of Falcon performance on the table.
Architecture-specific alpha calibration is therefore necessary when transferring \method to a new recurrent backbone.

\section{Parameter-Matched LoRA Per-Seed Results}
\label{app:matched_perseed}

Table~\ref{tab:lora_matched_perseed} reports
per-seed results for the higher-rank LoRA experiments. At rank~48, one seed
(555) collapses to 4.8\% accuracy; excluding this outlier, the mean is
$+7.3 \pm 5.5$~\pp, still far below \methodshort's $+23.6 \pm 1.7$~\pp. At
rank~64 (matched to \methodshort's 12.6M parameters), 8 of 10 seeds are
negative, with deltas ranging from $-41.7$~\pp{} to $+13.1$~\pp. All
configurations use identical training data, learning rate ($5 \times 10^{-4}$),
and steps (50); \texttt{lora\_alpha} is set to $2r$ following standard practice.

\begin{table}[h!]
\centering
\caption{Per-seed results for higher-rank LoRA on Qwen3.5-4B HumanEval.}
\label{tab:lora_matched_perseed}
\small
\begin{minipage}[t]{0.48\textwidth}
\centering
\textbf{Rank~48 (9.4M)}\\[4pt]
\begin{tabular}{@{}rrl@{}}
\toprule
\textbf{Seed} & \textbf{$\Delta$} & \textbf{Note} \\
\midrule
42   & $+6.0$~\pp  & \\
7    & $+10.7$~\pp & \\
123  & $+0.0$~\pp  & \\
256  & $+11.9$~\pp & \\
999  & $+3.6$~\pp  & \\
314  & $+2.4$~\pp  & \\
555  & $-44.0$~\pp & Collapsed \\
777  & $+3.6$~\pp  & \\
1337 & $+10.7$~\pp & \\
2024 & $+16.7$~\pp & \\
\midrule
\textbf{Mean} & $+2.1 \pm 17.0$~\pp & \\
\bottomrule
\end{tabular}
\end{minipage}%
\hfill
\begin{minipage}[t]{0.48\textwidth}
\centering
\textbf{Rank~64 (12.6M, matched)}\\[4pt]
\begin{tabular}{@{}rr@{}}
\toprule
\textbf{Seed} & \textbf{$\Delta$} \\
\midrule
42   & $-4.8$~\pp  \\
7    & $-33.3$~\pp \\
123  & $-29.8$~\pp \\
256  & $-41.7$~\pp \\
999  & $+10.7$~\pp \\
314  & $-19.0$~\pp \\
555  & $-6.0$~\pp  \\
777  & $-32.1$~\pp \\
1337 & $-11.9$~\pp \\
2024 & $+13.1$~\pp \\
\midrule
\textbf{Mean} & $-15.5 \pm 18.9$~\pp \\
\bottomrule
\end{tabular}
\end{minipage}
\end{table}

\section{\texorpdfstring{pass@$k$}{pass@k} Results}
\label{app:passk}

Table~\ref{tab:passk} reports aggregate pass@$k$ results.

\begin{table}[h!]
\centering
\caption{HumanEval pass@$k$ results on held-out problems 80--163 (3 seeds).}
\label{tab:passk}
\small
\begin{tabular}{@{}lccc@{}}
\toprule
\textbf{Method} & \textbf{pass@1} & \textbf{pass@5} & \textbf{pass@10} \\
\midrule
Baseline          & 41.9\% & 64.2\% & 69.4\% \\
LoRA (r24, 4.7M)  & 56.3\% & 64.1\% & 66.7\% \\
\methodshort (12.6M) & \textbf{70.3\%} & \textbf{84.5\%} & \textbf{88.5\%} \\
\bottomrule
\end{tabular}
\end{table}

\subsection{Per-Seed Results}
\label{app:passk_perseed}

Table~\ref{tab:passk_perseed} reports per-seed pass@$k$ estimates for all three
methods (3 seeds each). \methodshort exhibits low variance across seeds at
every $k$, while LoRA's pass@5 and pass@10 are within noise of the baseline.

\begin{table}[h!]
\centering
\caption{Per-seed HumanEval pass@$k$ results on held-out problems 80--163.}
\label{tab:passk_perseed}
\footnotesize
\begin{tabular}{@{}llccc@{}}
\toprule
\textbf{Method} & \textbf{Seed} & \textbf{pass@1} & \textbf{pass@5} & \textbf{pass@10} \\
\midrule
\multirow{3}{*}{Baseline}
 & 7   & 42.1\% & 63.6\% & 69.0\% \\
 & 42  & 43.5\% & 64.6\% & 70.2\% \\
 & 123 & 40.1\% & 64.3\% & 69.0\% \\
\midrule
\multirow{3}{*}{LoRA r24}
 & 7   & 56.4\% & 63.4\% & 65.5\% \\
 & 42  & 56.4\% & 64.7\% & 67.9\% \\
 & 123 & 56.0\% & 64.1\% & 66.7\% \\
\midrule
\multirow{3}{*}{\methodshort}
 & 7   & 69.6\% & 83.8\% & 86.9\% \\
 & 42  & 72.6\% & 86.4\% & 90.5\% \\
 & 123 & 68.6\% & 83.3\% & 88.1\% \\
\bottomrule
\end{tabular}
\end{table}

\section{Probing Protocol and Scaling Data}
\label{app:probing}
\label{app:scaling}

We probe whether the recurrent state encodes task-relevant information more strongly than the residual stream.
For each HumanEval problem, we extract the recurrent state $\state$ from every \gdn layer and the residual-stream hidden state, both at the final prompt token, projected to 64 dimensions via PCA (fit on train split only).
A logistic regression probe with 5-fold CV achieves $\AUC = 0.93$ from the recurrent state (best layer: 18) versus $\AUC = 0.90$ from the residual stream (best layer: 23).
The 2.5-point gap is directional evidence that recurrent states encode task-relevant information at least as strongly as residual-stream activations.

\vspace{4pt}
\noindent\textbf{Scaling data.}
HumanEval results across Qwen3.5 model sizes.
The 0.8B result is borderline ($p{=}0.076$); at 4B and above, gains are large and unambiguous.

\vspace{-2pt}
\begin{center}
\small
\begin{tabular}{@{}lrrccr@{}}
\toprule
\textbf{Model} & \textbf{GDN} & \textbf{State} & \textbf{Base} & \textbf{Tuned} & \textbf{$\Delta$} \\
\midrule
Qwen3.5-0.8B & 18 & 4.7M & 14.3\% & 16.9\% & $+2.6 \pm 3.7$~\pp \\
Qwen3.5-2B   & 18 & 4.7M & 16.7\% & 35.7\% & $+19.0 \pm 1.2$~\pp \\
Qwen3.5-4B   & 24 & 12.6M & 48.8\% & 72.2\% & $+23.6 \pm 1.7$~\pp \\
Qwen3.5-9B   & 24 & 12.6M & 32.1\% & 76.1\% & $+44.0 \pm 1.2$~\pp \\
\bottomrule
\end{tabular}
\end{center}

\end{document}

%% file: defs.tex
\newcommand{\method}{\texorpdfstring{S\textsubscript{0}~tuning}{S0 tuning}\xspace}
\newcommand{\methodshort}{\texorpdfstring{S\textsubscript{0}}{S0}\xspace}

\newcommand{\gdn}{GatedDeltaNet\xspace}
\newcommand{\falcon}{FalconH1\xspace}

\newcommand{\svec}{S_0}                    
\newcommand{\state}{S_t}                   
\newcommand{\stateshape}{\mathbb{R}^{H \times K \times V}}  

\DeclareMathOperator{\AUC}{AUC}

\newcommand{\pp}{\text{pp}\xspace}         

\newcommand{\mainresult}{+23.6}            
\newcommand{\loragap}{+10.8}              